VIETNAM NATIONAL UNIVERSITY – HOCHIMINH CITY
THE INTERNATIONAL UNIVERSITY
SCHOOL OF COMPUTER SCIENCE AND ENGINEERING

# IMPROVING THE PERFORMANCE OF THE K-MEANS ALGORITHM

By
Nguyen Tien Dung

Supervisor
Nguyen Duc Cuong, Ph.D.

A thesis submitted to the School of Computer Science and Engineering in partial fulfillment of the requirements for the degree of
Master of Information Technology Management

Ho Chi Minh city, Vietnam
October 2013

VIETNAM NATIONAL UNIVERSITY – HOCHIMINH CITY
THE INTERNATIONAL UNIVERSITY
SCHOOL OF COMPUTER SCIENCE AND ENGINEERING

# IMPROVING THE PERFORMANCE OF THE K-MEANS ALGORITHM

By
Nguyen Tien Dung

Supervisor
Nguyen Duc Cuong, Ph.D.

A thesis submitted to the School of Computer Science and Engineering in partial fulfillment of the requirements for the degree of
Master of Information Technology Management

Ho Chi Minh city, Vietnam
October 2013

**APPROVED BY SUPERVISOR**      **APPROVED BY COMMITTEE**

___________________________      ___________________________
Nguyen Duc Cuong, Ph.D.

___________________________

___________________________

___________________________

___________________________

THESIS COMMITTEE

# ACKNOWLEGMENTS

It is with deep gratitude and appreciation that I acknowledge the professional guidance of Dr. Nguyen Duc Cuong. This thesis has not been possible without his constant encouragements and supports during this challenging time.

My gratitude goes to the other lecturers of the school of Computer Science and Engineering, especially Dr. Pham Van Hau. His research experiences give me many treasured supports since the beginning of my studies.

# TABLE OF CONTENTS

# LIST OF TABLES

# LIST OF FIGURES

# ABBREVIATION

| | |
|---|---|
| KM | K-means Algorithm |
| IKM | Incremental K-means Algorithm |
| DKM | Divisive K-means Algorithm |
| 2PK-means | Two-Phase K-means |
| MPI | Message Passing Interface |
| PVM | Parallel Virtual Machine |
| Par2PK-means | Parallel Two-Phase K-means |

# ABSTRACT

The Incremental K-means (IKM), an improved version of K-means (KM), was introduced to improve the clustering quality of KM significantly. However, the speed of IKM is slower than KM. My thesis proposes two algorithms to speed up IKM while remaining the quality of its clustering result approximately. The first algorithm, called Divisive K-means, improves the speed of IKM by speeding up its splitting process of clusters. Testing with UCI Machine Learning data sets, the new algorithm achieves the empirically global optimum as IKM and has a lower complexity, $O(k*\log_2 k*n)$, than IKM, $O(k^2 n)$,. The second algorithm, called Parallel Two-Phase K-means (Par2PK-means), parallelizes IKM by employing the model of Two-Phase K-means. Testing with large data sets, this algorithm attains a good speedup ratio, closing to the linearly speed-up ratio.

# CHAPTER I

# INTRODUCTION

## 1.1 Background

K-means (KM) [1] is one of the most widely used algorithms in Data Mining because of its simplicity and computational efficiency. However, K-means has some known drawbacks that have been studied by several researchers. One of K-means drawbacks is its convergence to local optimums, due to its random initialization of cluster centers. In 2004, the Incremental K-means algorithm (IKM) [2] was introduced to overcome this problem. This algorithm can achieve the empirical global optimum regardless the random initialization of cluster centers. However, its level of complexity ($O(K^2n)$) is higher than KM ($O(Kn)$).

There are two methods to speed up IKM while maintaining the quality of its clustering result:

- Speeding up the stepping process of the number of clusters of IKM
- Applying a parallel strategy to take the advantage of multiple computing units

The IKM uses the strategy of stepping the number of clusters by 1, from 1 to K. It means that IKM has to execute K-means $K$ times. Therefore, if the speed of the increasing process of the cluster number is improved, the speed of IKM will be improved. To speed up this stepping process, a binary splitting strategy can be used to double the current number of clusters until it reaches the required *K*. This greedy splitting process can speed up the number of cluster exponentially, but it also introduces a significant decrease in the clustering performance because a new cluster can be inserted to an inappropriate position. By making a combination of the binary splitting of clusters and several adjustments, an improved version of IKM will run faster than IKM, and approximately achieves the empirical global optimum of IKM.

The Two-Phase K-Means (2PK-means) [3] is introduced to scale up K-means to process large data sets. By employing the strategy of dividing the clustering process into two phases, 2PK-means only requires one scan over a large data set. In contrast, many available parallel versions of K-means algorithm require many scans over a data set until the clusters are stable. In case of processing a large data set (up to several TBs), scanning a data set one time will help to reduce the communication cost between many computing units in a parallel system. In addition to this, the Phase 1 of 2PK-means applies KM on each data segment independently (each data segment has a number of data objects), so this phase can be parallelized to reduce the computational cost. In general, using the model of 2PK-means will be an appropriate parallel strategy to improve the speed of IKM.

**1.2     Objectives**

The objectives of my thesis are:

- Proposing an improved version of the Incremental K-means algorithm to reduce its complexity while maintaining the quality of its clustering result.
- Proposing a parallel version of the Incremental K-means algorithm by employing the model of Two-Phase K-means to take the advantage of multiple computing units.

**1.3     Thesis Structure**

This dissertation is divided into 4 chapters. Chapter I, ***Introduction***, offers the problem statement, the objectives and the structure of my thesis. Chapter II, ***Literature Review***, concentrates on reviewing three approaches to improve the performance of K-means algorithm, and the limitations of previous parallel versions of K-means algorithm. Chapter III, ***Divisive K-means Algorithm***, describes the Divisive K-means algorithm, and evaluates its performance in comparison with the original version of K-means and Incremental K-means. Chapter IV, ***Parallel Two-Phase K-means***, describes the Parallel Two-Phase K-means algorithm, and evaluates its performance. Finally, Chapter V, ***Conclusion and Discussion***, discuss about the results as well as the potential research trends.

# CHAPTER II

# LITERATURE REVIEW

This chapter is divided into five sections. Section 2.1, ***K-means algorithm***, discusses about the original version of K-means algorithm. Section 2.2, ***Three approaches to overcome the convergence to local optimum of K-means***, discusses about the related works to improve the performance of the original K-means algorithm. Section 2.3, ***Incremental K-means algorithm***, discusses about Incremental K-means algorithm. Section 2.4, ***Parallelizing the K-means algorithm***, discusses about the limitations of previous parallel versions of K-means algorithm. Section 2.5, ***Two-Phase K-means algorithm***, discusses about the main ideas of Two-Phase K-means algorithm to handle a large data set.

## 2.1 K-means Algorithm

K-means (KM) [1] is one of the most widely used algorithms for Data Mining because of its efficiency and low complexity, $O(Knl)$, where n is the number of objects, K is the required number of clusters and l is the maximum number of iterations.

One of K-means drawbacks is its convergence to local optimums due to its random initialization of cluster centers. Three approaches to overcome this problem are reviewed in the next section.

## 2.2 Three Approaches To Overcome The Convergence To Local Optimums Of K-means

A drawback of K-means algorithm is that it often converges to local optimums. To overcome this problem, improved studies can be classified into three main approaches as follows:

- Finding a good initialization of cluster centers.
- Initializing randomly cluster centers with a modified learning mechanism.
- Stepping the number of clusters from 1 to the expected number of clusters $K$ with a mechanism to initialize new clusters.

In the first approach, there have been several attempts to find a good initialization for cluster centers in the K-means algorithm. Better-initialized positions can be found by searching possible positions based on a heuristic [4] or the convex hull characteristic [5]. Cluster centers can also be initialized by using a kd-tree to perform a density estimation of the data [6]. Another solution is to assume that "each of the attributes of the pattern space are normally distributed", cluster centers based on seed points that are calculated based the mean and standard deviation of data attributes [7]. In spite of the improvements in performance, these initialized methods have not achieved a satisfactory result and introduced additional complexity.

In the second approach, there are several studies in modifying the learning strategy of K-means to overcome the problem of local optimum. After initializing cluster centers, to improve its performance with a modified learning mechanism, Fritzke [8] suggested a new jumping operation to facilitate the algorithm's convergence and assist it in escaping from local minima. In the same direction as Fritzke's work, the utility index is used as a reference [9]. Chinrungrueng and Sequin [10] proposed a new updating method introducing a restriction hypothesis for the problem's underlying object distribution. The stochastic relaxation scheme was applied to the K-means method to improve its performance [11]. However, the proposed algorithms, which use this approach, have failed to reach the global optimum.

In the third approach, instead of initializing $K$ clusters at the beginning, only one cluster is initialized, and then the number of clusters is stepped by 1, from 1 to $K$. In each stepping, a new cluster is inserted to the current cluster set. There are several methods to insert the new cluster. In Global K-means [12], the center of the new inserted cluster is globally searched for possible data points. This searching step helps the Global K-means finding a good position to initialize the new cluster but it requires a higher level of complexity. An improved version of Global K-means [13], that computes the center of the new inserted cluster by minimizing an auxiliary cluster function, can achieve a better result, but its complexity is even higher than the Global K-means. Another improvement of Global K-means [14] initializes the new cluster in a position that minimizes an introduced heuristic function. In the Bisecting K-means algorithm [15], instead of stepping the number of clusters during the learning process, the current number of clusters is doubled by splitting each cluster into two clusters. This splitting can speed up the incrementing process but its naïve division can put a new cluster center into sub-optimal positions. When the required number of clusters is not in a form of $2^n$, the algorithm does not  mention how to achieve the solution.

The next section discussed about the Incremental K-means Algorithm, which employs the third approach – stepping the number of clusters from 1 to K.

**2.3    Incremental K-means Algorithm**

The Incremental K-means algorithm (IKM) [2] uses the strategy of stepping the number of clusters from 1 to $K$. The main idea of IKM algorithm is described in Figure 1. In IKM, the original K-means is executed $K$ times, with the current number of clusters $K_c$ being stepped by 1 as in Step 2. With the heuristic of selecting a good position for the new added cluster, this new cluster is inserted into the cluster of the cluster set with largest distortion error. Therefore, IKM can achieve the empirical global optimum on data sets with numerous attributes. However, the complexity of IKM algorithm is higher (O ($K^2n$)), in comparison with KM (O($Kn$)).

> *Incremental K-means*
>
> 1. Step 1 – "Initialization":
>    Assign $K_c=1$
>
>    *K-means_Learning($K_c$)*
>
> 2. Step 2 – "Stepping $K_c$":
>    While ($K_c < K$)
>
>        Increase $K_c$ by 1.
>
>        Insert a new cluster to the cluster with the largest distortion
>
>        *K-means_Learning($K_c$)*
>
>    End
>
> *K-means_Learning($K_c$)*
>
> 1. If the cluster set does not move, stop.
> 2. For each data object in the data set
>        Assign the data object to the nearest cluster, update that cluster's information
> 3. Go to step 1

FIGURE 1. The Incremental K-means Algorithm (IKM) (Pham, Dimov, & Nguyen, 2004)

## 2.4 Parallelizing The K-means Algorithm

By employing different parallel strategies, the K-means algorithms will reduce computational time by taking the advantage of multiple computing units. This section reviews many different parallel versions of the K-means algorithm. Particularly, there are several parallel versions of the K-means algorithm implemented on different parallel programming frameworks, such as Parallel Virtual Machine (PVM), Message Passing Interface (MPI) or MapReduce. These parallel versions utilize the computing power of multiple computing nodes to speed up the clustering process of K-means.

The parallel K-means of Kantabutra and Couch [16] is implemented in the master/slave model on the Message-Passing Interface (MPI) framework and executed on a network of workstations. This algorithm uses one slave to store all data objects of a cluster. It divides K subsets of the data set and sends each subset to a slave. In each iteration of K-means, a new center is re-calculated in each slave and then broadcasted to other slaves. After that, data are sent between slaves to guarantee that, each subset on a slave only keeps data objects nearest to the center in that slave. However, the data re-arrangement requires a big data transmission between slaves and makes this strategy not applicable for big data sets.

The parallel K-means algorithm of Zhang et al. [17] is realized in the master/slave model based on the PVM framework and executed on a network of workstations. In the early state of the algorithm, the master reads the data and randomly initializes the cluster set. In each iteration, the master sends the cluster set to all slave nodes. The master divides the data set into S subsets (S can be larger than K), and consequently sends each subset to a slave node. Each slave receives a subset of data, and clusters independently this subset based on the cluster set and then sends its intermediate result back to the master node. The master node re-computes the values of the cluster centers based on the intermediate results received from slaves, and then the next iteration is executed. This process is repeated until the cluster set is stable. This parallel version of K-means requires that the data set is completely loaded on the master node and a synchronization of data is processed at the end of each iteration.

The parallel K-means algorithm of Tian et al. [18] shares the same strategy as the one of Zhang et al. [17]. It requires that the data set is completely loaded on the master node and divided into m subset (m is the number of processors). The paper only estimates the complexity of the proposed algorithm. There is no practical implementation to evaluate the empirical performance of the proposed algorithm.

A parallel K-means algorithm, called ParaKMeans [19], is implemented to cluster biological genes in the multi- threading approach (on a single computer). The parallel model in this algorithm is similar to the Tian's algorithm. The differences are the measurement of the cluster's quality and the stop condition.

A distributed K-means algorithm is introduced in [20]. It is designed to execute on multi- processor computers. Randomly spliting data subset is delivered to each processor before starting the algorithm. In the beginning, each processor randomly initializes the center of its K cluster centers. In each iteration, the cluster set of each processor is re-calculated based on its data subset. After that, its cluster set is broadcasted to other processors, and then its cluster set is re-calculated again based on the received data. The process is repeated until the cluster set is stable. In general, the parallel strategy of this paper is similar to Zhang's algorithm except using the master node, so that it has the same drawbacks.

With the introduction of MapReduce in 2004 [21], all following parallel versions of K- means in this section are implemented in this framework. The MapReduce framework uses a machine to play the role of the master node in the master/slave model. The Master node splits data into subsets, sends each subset to each Mapper (playing a role of a slave node), invokes the action of all Mappers. Each Mapper processes its data subset to create intermediate data and then sends its intermediate data to Reducers. Reducers collect intermediate data from Mappers to create final results. A node (Master, Mapper, Reducer) can be an independent computer.

Chu's work [22] proposed a framework that applied the parallel programming method of MapReduce on several Machine Learning algorithms, including K-means. In the parallel K-means algorithm in this framework, each Mapper works on different data segments, and then its intermediate data (the sum of vectors in each data subgroup) is sent back to the Master node. After collecting all intermediate data from Mappers, the Master sends the received data to a Reducer to compute the new centroids, and then sends the final clustering result back to the Master. In comparison with the original K-means algorithm, the repeated converged process is not mentioned in the parallel version in this framework, so that the parallel version cannot reach to a good clustering result.

Another parallel K-means algorithm, called PKMeans, is introduced in [23] and implemented on MapReduce. This algorithm is also used in [24] for document clustering. Another type of nodes, called Combiner, is used in PKMeans. In each iteration, the Mapper, Combiner and Reducer are serially executed. A Mapper only assigns each sample to the nearest center. A Combiner, which is executed on the same computer with the Mapper, partially sums the values of the data points assigned to the same cluster and then returns an array of records. Each record stores the sum of values and the number of data points of a cluster. This array will be sent to the Reducer to compute the new position of cluster centers, which is a global variable and can be accessed by Mappers. Several iterations are repeated until the cluster set is stable. Therefore, the model and the limits of this work are similar to Zhang's algorithm.

In conclusion, several parallel versions of K-means employ many data parallel strategies. They are different in using or not using MapReduce framework. In case of not using MapReduce, each slave node uses the initialized data from the master node, processes its sub data set and then synchronizes local data by sending them back to the master node or broadcasting them to other computing nodes before the next iteration. This strategy has several drawbacks, such as the master node has to load the full data set before delivering data to computing nodes, a synchronization step is required in the end of each iteration, several scans over the data set is also required. When using MapReduce, a Mapper processes each data object at a time, so that a Mapper is called several times. Because the algorithm requires several iterations, the communication cost between these nodes is high.

## 2.5 Two-Phase K-means Algorithm

The Two-Phase K-Means (2PK-means) [3] is introduced to scale up K-means to process large data sets. The K-means algorithm requires several scans over data sets, so that, to speed up the data accessing, the data set has to be fully loaded to the computer memory. With large data sets (up to several TBs), this requirement is hard to fulfill. 2PK-means is introduced to overcome this drawback. 2PK-means has 2 phases. In Phase 1, 2PK-means sequentially loads and processes all data segments of the data set (each data segment contains a number of data objects) to produce the temporary cluster set which is stored for Phase 2. The K-means algorithm is used in Phase 1 due to its low complexity. In Phase 2, 2PK-means clusters all the intermediate data to create final clustering result by IKM.

With the strategy of dividing the clustering process into two phases, 2PK-means only requires one scan over the large data set. It means that computers with limited memory can apply this strategy. The clustering result of 2PK-means is approximately the same with KM [3]. If 2PK-means uses IKM in both phases, 2PK-means can approximately achieve the same clustering result with IKM when working on the whole data set.

In 2PK-means, the phase 1 loads and processes all data segments of the data set independently. Therefore, parallelizing the phase 1 of 2PK-means will increase the speed of 2PK-means, while it still remains the clustering result of 2PK-means unchanged. This is the main idea of the new proposed algorithm, called Parallel Two-Phase K-means. This algorithm applies IKM in two phases of 2PK-means, and parallelizing the phase 1 of 2PK-means. Chapter IV discusses the details of this algorithm, as well as its performance evaluation.

# CHAPTER III

# DIVISIVE K-MEANS

## 3.1 Algorithm Description

The Divisive K-means algorithm (DKM) combines the fast splitting process of the Bisecting K-means algorithm and the heuristically insertion of a new cluster of IKM. The algorithm of DKM is described in Figure 2. DKM starts with $K_c$ (the current number of clusters) equal to 1. In Step 2, $K_c$ is doubled by splitting clusters, and then the learning process of function *K-means_Learning($K_c$)* will update the position of clusters. Step 2 is repeated until $K_c$ is smaller than $K$ by margin $K_t$. After the bisecting process (Step 2) stops, a stepping process (Step 3) is used to step $K_c$ to $K$.

*Divisive K-means*

1. Step 1 – "Initialization":
   Assign $K_c = 1$

   *K-means_Learning($K_c$)*

2. Step 2 – "Doubling $K_c$":
   while ($2 * K_c \leq K - K_t$)

   Split each cluster into 2 clusters

   *K-means_Learning($K_c$)*

   $K_c \leftarrow 2 * K_c$

   end while

   if ($K_c < (K - K_t)$)

   Split each cluster of ($K - K_t - K_c$) clusters with the largest distortion into 2 clusters

   *K-means_Learning($K_c$)*

   $K_c \leftarrow K - K_t$

   end if

3. Step 3 – "Stepping $K_c$":
   while ($K_c < K$)

   Increase $K_c$ by 1.

   Insert a new cluster to the cluster with the largest distortion

   *K-means_Learning($K_c$)*

   end while

*K-means_Learning($K_c$)*

1. If the cluster set does not move, stop.
2. For each data object in the data set
   Assign the data object to the nearest clusters; update that cluster's information

3. Go to step 1

Notes:

$$K_t = \begin{cases} K & \text{if } k < 7 \\ \min(\max(3, K * 10\%), 5) & \text{if } k \geq 7 \end{cases}$$

FIGURE 2. The Divisive K-means Algorithm (DKM)

In the last bisecting step of DKM, the increased number of clusters is often smaller than the current number of clusters. In this case, instead of splitting all current clusters, new clusters are inserted into existing clusters with the largest distortions.

It is possible that the new clusters can be inserted into a small-distortion-error cluster. Consequently, the clustering performance of DKM can be reduced significantly. Step 3 of DKM uses the same mechanism as IKM by stepping $K_c$ from $(K - K_t)$ to $K$. The selection of $K_t$ determines the performance of DKM. A larger $K_t$ results in a slower DKM and a higher probability that DKM can achieve a smaller distortion error. In our practical experience, $K$ can be selected from 3 to 5 and relative to the number of required clusters. Specially, when $K$ is smaller than 7, the number of divisions in Step 2 is small so the acceleration of DKM is also small. Therefore, with such $K$, $K_t$ is set as $K$ so Step 2 is bypassed and DKM becomes IKM.

The complexity of DKM is calculated in cases of $K$ larger than 7 by the formula $O(K*(log_2K+K_t)*n*number\_of\_iterations)$, where n is the number of objects, and the number_of_iterations is the largest possible number of iterations in function *K-means_Learning* in Figure 2. When the number of required clusters $K$ is large for very large data sets, $K_t$ is considerably smaller than $K$, so the complexity can be reduced to $O(K*log_2K*n*number\_of\_iterations)$. Compared with the complexity $O(K*n*number\_of\_iterations)$ of the K-means algorithm, DKM requires $log_2K$ times more iterations. Compared with the complexity $O(K^2*n*number\_of\_iterations)$ of IKM, DKM requires $K/log_2K$ times less iterations.

The algorithm DKM uses the doubling strategy in Step 2 to speed up the process of increasing the number of clusters from 1 to close to $K$. Step 2 cannot be used to increase the number of clusters to $K$ because the clustering performance can be reduced by this greedy splitting. Step 3 of DKM is introduced to step the number of clusters from ($K$-$K_t$) to $K$ to guarantee a good clustering performance. In general, DKM will be faster than IKM but still achieve the empirical optimum as IKM

## 3.2 Performance Evaluation

Three algorithms IKM, DKM and the original K-means were implemented in a Java-based machine learning software workbench called Weka [25], and executed on a personal computer with a CPU Intel core i5 and 4GB memory. Eight real data sets from the UCI Repository [26] were used to test the performances of the new proposed algorithm (DKM), IKM and the original K-means. These data sets were selected because they consist of only numerical attributes. In each downloaded data set, the class attribute was deleted. The characteristics of these data sets are represented in Table 1.

TABLE 1. The Characteristics of Tested Data Sets

| Data sets | Number of attributes | Number of objects |
|---|---|---|
| Balance-Scale | 4 | 635 |
| Breast Cancer | 32 | 569 |
| Ionosphere | 34 | 351 |
| Iris | 4 | 150 |
| Letter | 16 | 20000 |
| Pima | 8 | 768 |
| Wine | 13 | 178 |
| Zoo | 17 | 101 |

An algorithm in the K-means family can be stopped by the specification of termination conditions, such as a predefined number of iterations, and the total displacement of cluster centers after an iteration is smaller than a given value ε. In the experiments, these two termination criteria were used for the function K-means_Learning (see Figure 2). Particularly, the maximum number of iterations was empirically set to 20 and ε to $10^{-7}$. The algorithm stopped when one of these conditions were satisfied.

All three versions of K-means (original K-means, IKM and DKM) were executed 100 times for each given *K*. Figure 3 shows the results obtained by applying three different versions of the K-means algorithm to the 8 selected data sets. $I_a$ and $I_{min}$ are the average and minimum distortion values of cluster distortion errors. $I_a$ was calculated for each algorithm with each given *K*. $I_{min}$ was calculated based on the results of all three tested algorithms for each given *K*. The ratio $I_a/I_{min}$ represents the performance of the algorithm. When this ratio approaches to 1 with a given *K*, it means that the tested algorithm frequently achieved the empirical optimum with that *K*.

Figure 3 shows that, for all tested data sets with all given *K*, the ratios $I_a/I_{min}$ of IKM and DKM approach to 1. This means that IKM and DKM do not depend on the specific characteristics of the data sets and the value *K*, and produce reliable and optimal clustering of objects. The ratios $I_a/I_{min}$ of the original K-means with different *K* is always much larger than 1. It means that the original K-means rarely approaches the global optimum.

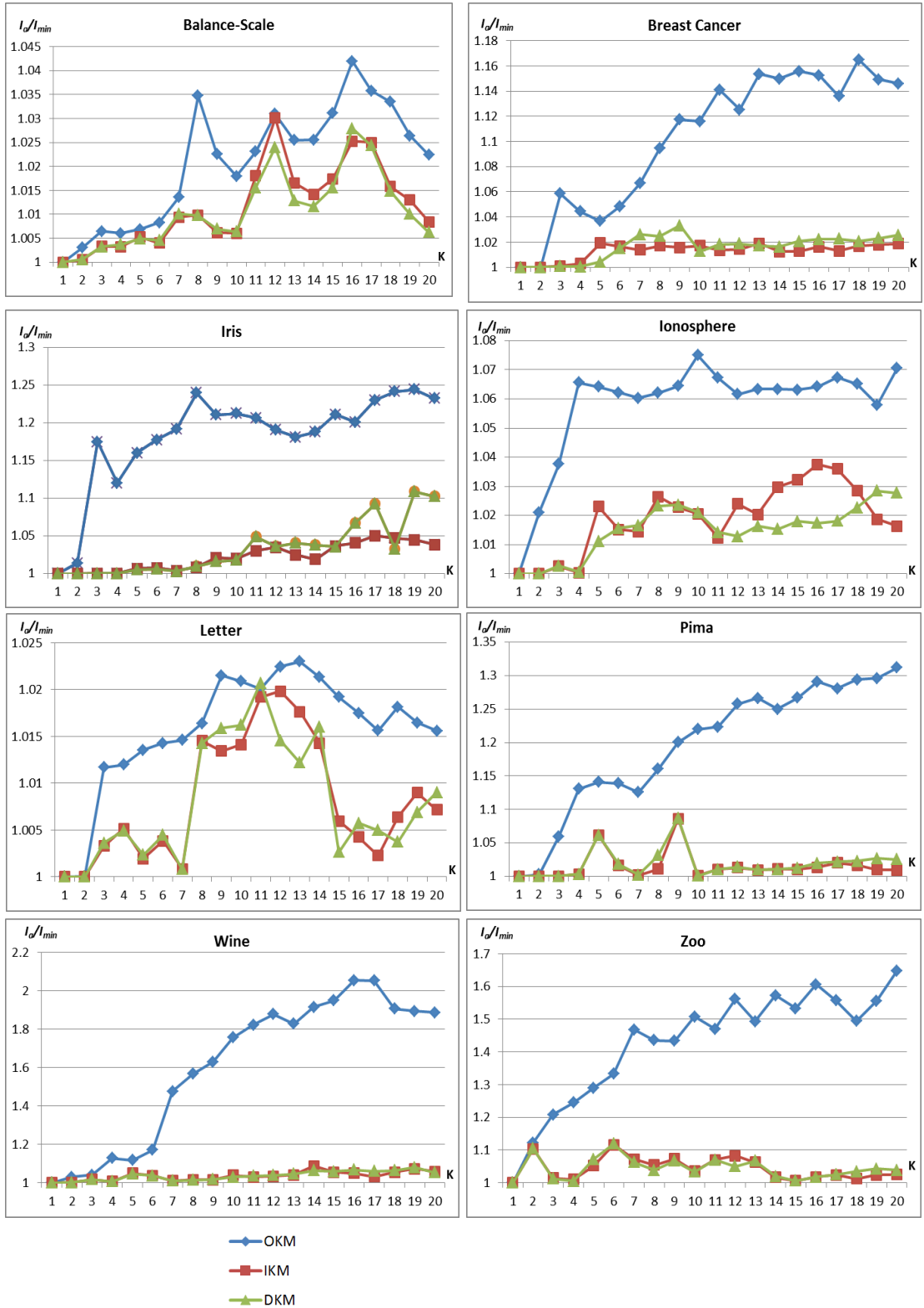

FIGURE 3. Clustering Results of Original K-means, Incremental K-means

and Divisive K-means.

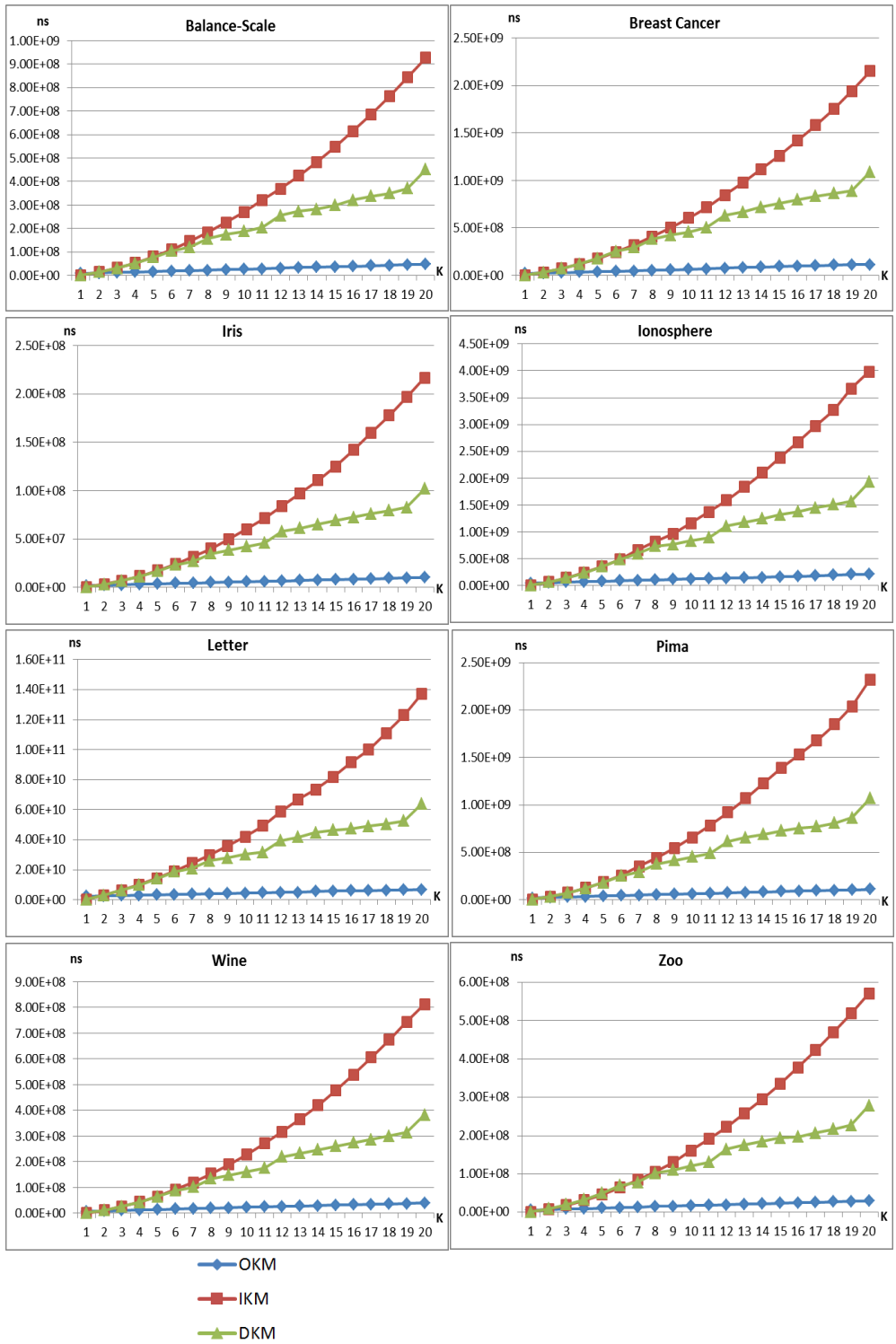

FIGURE 4. The Running Time of K-means, Incremental K-means

and Divisive K-means.

Figure 4 shows the running time of the original K-means, IKM and DKM. When $K$ is smaller than 7, IKM and DKM have the same running time. With $K$ larger than 7, DKM has a smaller running time, similar to the result in the calculation of its complexity.

In Figure 4, the running time of DKM for all tested data sets increases linearly with $K$ from 12 to 19 and has a small step with $K$ equal to 20. This observation can be explained by calculating the number of calls of function *K-means_Learning* in DKM. For example, when $K$ equal to 19, function *K-means_Learning* is called 7 times with parameter $K_c$ as 2, 4, 8, 16, 17, 18, and 19. For another example, when $K$ equal to 20, function *K-means_Learning* is called 8 times with parameter $K_c$ as 2, 4, 8, 16, 17, 18, 19, and 20. Therefore, function *K-means-Learning* is called in the case of $K$ equal to 20 one more time than the case of $K$ equal to 19. This makes a small step in the running time in Figure 4. When $K$ is larger than 20 and smaller than 36, the running time will increase linearly again.

# CHAPTER IV

# PARALLEL TWO-PHASE K-MEANS

## 4.1 Algorithm Description

A parallel version of IKM, called Parallel Two-Phase K-means (Par2PK-means), is developed based on the model of 2PK-means and implemented on Hadoop MapReduce framework. The algorithm of Par2PK-means is described in Figure 5. When there is an available Mapper, it reads a data segment and executes the IKM algorithm on that data segment to create an intermediate clustering result. Reducer retrieves the intermediate clustering result of all Mappers. When all Mappers finish their tasks, the Master invokes a Reducer to execute IKM on all received intermediate results from the Mappers to create the final clustering result.

Parameter $K_t$ decides the speed and the clustering quality of Par2K-means. Phase 1 of Par2PK-means plays a role of a compressor that reduces the number of data objects in a data buffer to $K_t$ clusters. Parameter $K_t$ decides the size of the intermediate data, which is processed in Phase 2. So that $K_t$ controls the speed of the algorithm. The smaller $K_t$ is, the faster Phase 1 of Par2K-means is. However, $K_t$ cannot be too small to maintain the final clustering quality. The larger $K_t$ is, the higher the quality of clustering result is [20]. Besides, Par2K- means employs the model of Two-Phase K-means, so that it produces an approximate result of the sequential version of Two-Phase K-means. Therefore, the selection of $K_t$ is trade-off between the speed and the clustering quality. In Two-Phase K-means, $K_t$ is selected correspondent to the size of a data buffer.

> ***Parallel Two-Phase K-means:***
>
> Input: $K$ and $K_t$ being the expected number of clusters and the temporary number of clusters for Mappers
>
> $L$ being the length of the data segment
>
> Output: the cluster set
>
> Algorithm:
>
> ***Mapper:*** map (*inputValue, inputKey, outputValue, outputKey*)
>
> 1. Load a data segment from *inputValue*
> 2. Execute IKM($K_t$) on the loaded data segment
> 3. Format *outputValue* as the clustering result (cluster centers with number of belonged number of data objects)
> 4. Output (*outputValue, outputKey*)
>
> Note: *outputKey* has the same value for all Mappers
>
> ***Reducer:*** reduce (*inputValue, inputKey, outputValue, outputKey*)
>
> 1. Load intermediate result from *inputValue*
> 2. Execute IKM($K$) on the received intermediate result
> 3. Output *the cluster set* as the final result

FIGURE 5. The Parallel Two-Phase K-means Algorithm

### 4.2 Performance Evaluation

The proposed algorithm is evaluated its speed-up ratio. The experiments are conducted on a virtual computer cluster powered by Openstack cloud computing infrastructure. Each virtual computer has a 2.8GHz CPU and 1GB of memory. Hadoop version 1.0.4 and Java 1.6.0_26 are used as the MapReduce system for all experiments.

The experiments use data set CoverType from the UCI Repository [26]. This data set has 581,012 data objects with 52 attributes. In these experiments, only the first 4 attributes are used. The data set CoverType is enlarged by adding noise to create two tested data sets. The enlarged data set 1 has 2,324,048 data objects, 4 attributes, and a size of 87.6MB. The enlarged data set 2 has 29,050,600 data objects, 4 attributes, and a size of 1.23GB.

The algorithm is tested with different the number of slave nodes (1, 2, 4 and 8). The number of data objects in one data segment is varied from 0.1% to 1% the size of data sets. Number of clusters K and $K_t$ are selected as 10. The speed-up ratios of Par2PK-means in different number of slave nodes are compared to the linear ratio.

The evaluation results are shown in Table 2 and Figure 6. In the experiment with data set 1 (87.6MB), the number of data objects in one data segment is selected as 1% the size of the data set 1. It can be seen that, the speedup ratio of Par2PK-means approaches its limit (linear speedup). The lost percentage of speedup ratio of Par2PK-means compared to the linear ratio is around 10% due to the initialization of Hadoop. In the experiment with the data set 2 (1.23GB), the number of data objects in one segment is selected as 0.1% of the data size. In this case, it can be seen that the computational cost of a MapReduce job is larger when the data size is larger. Therefore, the ratio of time for the MapReduce tasks' initialization in total running time is reduced as the data size is increased. This reduction means that its speedup ratio is closer to the linear ratio (a perfect situation for parallelizing an algorithm).

TABLE 2. Speedup Ratio of the Par2PK-means in Different Data Sets

| Number of Computing Nodes | Speedup Ratio | |
|---|---|---|
| | Data Set 1 - 87.6MB | Data Set 2 - 1.23GB |
| 1 | 1 | 1 |
| 2 | 1.95 | 1.99 |
| 4 | 3.75 | 3.96 |
| 8 | 6.98 | 7.77 |

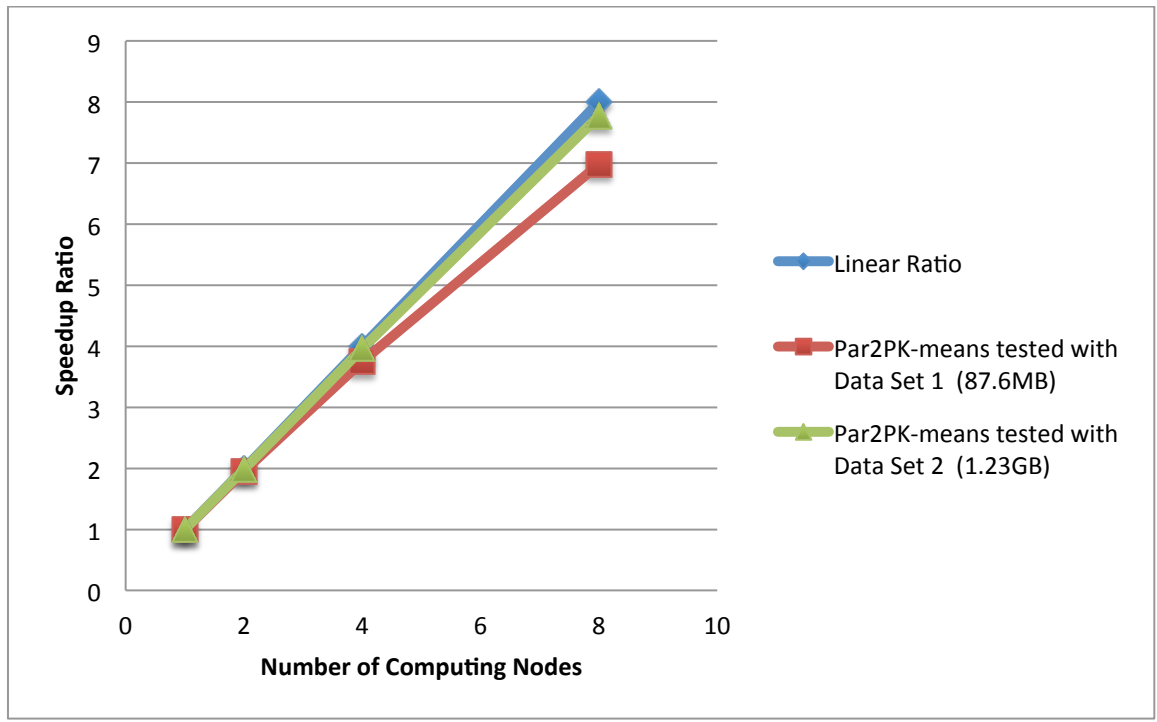

FIGURE 6. The Speed-up Ration of Different Data Sets

# CHAPTER V

# CONCLUSION AND DISCUSSION

This thesis has introduced an improved version of IKM, and a parallel version of IKM on the model of Two-Phase K-means algorithm.

The Divisive K-means algorithm (DKM) has been tested on eight real data sets. The algorithm consistently outperforms the original K-means algorithm and achieves the empirical global optimum as the Incremental K-means algorithm (IKM). Compared with IKM, the running time of DKM is significantly reduced, and its clustering quality is approximately the same. However, DKM still exists two gaps. Firstly, the function of $K_t$ should be studied further to optimize the clustering quality as well as the running time of DKM. Secondly, DKM does not outperform IKM and K-means in term of clustering result at several cases. The explanations and solutions for these problems should be studied in the future.

The Parallel Two-Phase K-means algorithm has achieved a good speedup ratio on tested data sets. However, its performance can be analyzed with other parallel strategies. Its incremental attribute (the algorithm can stopped on a number of data segments and give the best so-far clustering result) should be studied.

# LIST OF REFERENCES

# Appendix A – Publication

During my thesis, I have a scientific paper that is accepted in the 13th International Conference on Computational Science and Its Applications (ICCSA 2013). This paper introduces a new algorithm called Parallel Two-Phase K-means, which is discussed in Chapter IV. The full paper is enclosed in this appendix.

# Parallel Two-Phase K-Means[*]

Cuong Duc Nguyen, Dung Tien Nguyen, and Van-Hau Pham

International University – VNU-HCM
{ndcuong,ntdung,pvhau}@hcmiu.edu.vn

**Abstract.** In this paper, a new parallel version of Two-Phase K-means, called Parallel Two-Phase K-means (Par2PK-means), is introduced to overcome limits of available parallel versions. Par2PK-means is developed and executed on the MapReduce framework. It is divided into two phases. In the first phase, Mappers independently work on data segments to create an intermediate data. In the second phase, the intermediate data collected from Mappers are clustered by the Reducer to create the final clustering result. Testing on large data sets, the newly proposed algorithm attained a good speedup ratio, closing to the linearly speed-up ratio, when comparing to the sequential version Two-Phase K-means.

**Keywords:** Data Clustering, K-means, Parallel Distributed Computing, MapReduce.

## 1 Introduction

The K-means algorithm is one of the most popular Data Mining algorithm. There are several parallel versions of the algorithm implemented on different programming frameworks, such as Parallel Virtual Machine (PVM), Message Passing Interface (MPI) or MapReduce. These parallel versions utilize the computing power of slave nodes to speed up the clustering process of K-means.

There are two drawbacks can be recognized in the current parallel versions of the K-means algorithm. Firstly, in some parallel algorithms [1-4], data are divided into equal subsets to compute on slave nodes. This assumes that the processing times on slave nodes are equivalent. In these algorithms, intermediate data are collected from slave nodes to update the global information by a Master node and then to broadcast this global data to slave nodes. Therefore, a synchronization is required the end of each iteration. Secondly, several parallel versions [1-5] require the data set is fully loaded, divided and sent to slave nodes. If the memory size of a slave node is smaller than the size of the data subset, the data will be swapped between memory and hard disk, so that this swapping will slow down the algorithm.

To overcome the limitations mentioned above, this paper introduces a new parallel version of K-means, called Par2PK-means, that is implemented on MapReduce. Par2PK-means reads the data set as a stream of data segments, independently processes each data segment on slave nodes to create intermediate data and finally processes intermediate data to create final clustering result. In Par2PK-means, each data segment is independently processed once on a Mapper so that the communication cost between Mappers and Reducer are reduced.

---

[*] The work is supported by DOST, Hochiminh City under the contract number 283/2012/HD-SKHCN.

The remainder of the paper is organized as fellows. Section 2 aims at discussing about the related works. The new algorithm is described in Section 3. Evaluation of the new parallel version is presented in Section 4. Section 5 concludes the paper

## 2    Related Works

### 2.1    The K-Means Algorithm

K-means (KM) was first introduced by MacQueen in 1967 [6], and it has become one of the most widely used algorithms for Data Mining because of its efficiency and low complexity, O($Knl$), where $n$ is the number of objects, $K$ is the required number of clusters and $l$ is the maximum number of iterations. However, K-means is often converge to a local optimum. To overcome this drawback, the Incremental K-means algorithm (IKM) algorithm [7], an improved version of K-means, can empirically reach to the global optimum by stepping $k$ from 1 to the required number of clusters. However, IKM has a higher complexity, O($K^2nl$).

The Two-Phase K-Means (2PKM) [8], is introduced to scale up K-means to process large data sets. The K-means algorithm requires several scans over data sets, so that, to speed up the data accessing, the data set has to be fully loaded to the computer memory. With large data sets having several TBs, this requirement is hard to fulfill. 2PKM is introduced to overcome this drawback. 2PKM has 2 phases. In Phase 1, 2PKM loads and processes piece by piece of the dataset to produce the temporary cluster set which is stored for Phase 2. The K-means algorithm is used in Phase 1 due to its low complexity. In Phase 2, 2PKM clusters all the intermediate data to create final clustering result by IKM.

With the strategy of dividing the clustering process into two phases, 2PKM only requires one scan over the large data set and particularly this can be done by computers with limited memory. It can achieve approximate clustering result of the result that is created by K-means working the whole data set [8]. If 2PKM uses IKM in both phases, 2PKM can achieve approximate clustering result of the result that is created by IKM working the whole data set. Therefore, in this paper, IKM is used in both two phases of 2PKM.

### 2.2    Parallelizing the K-Means Algorithm

The parallel K-means of Kantabutra and Couch [5] is implemented in the master/slave model on the Message-Passing Interface (MPI) framework and executed on a network of workstations. This algorithm uses one slave to store all data objects of a cluster. It divides $K$ subsets of the data set and sends each subset to a slave. In each iteration of K-means, a new center is re-calculated in each slave and then broadcasted to other slaves. After this center broadcasting, data are sent between slaves to make each subset on a slave only keeps data objects nearest to the center in that slave. This step of data re-arrangement requires a big data transmission between slaves and makes this strategy not suitable for big data sets.

The parallel K-means algorithm of Zhang et al. [1] is realized in the master/slave model based on the PVM framework and executed on a network of workstations. In the early state of the algorithm, the master reads the data and randomly initialize the cluster set. In each iteration, the master sends the cluster set to all slave nodes. The master divides the data set into $S$ subsets ($S$ can be larger than $K$) and consequently sends each subset to a slave node. A slave receives a subset of data, independently clusters this subset based on the cluster set and then sends its intermediate result back to the master node. The master node re-computes the position of the cluster centers based on the intermediate results received from slaves and then to starts a new iteration until the cluster set is stable. This parallel version of K-means requires the full load of the data set on the master node and a synchronization of data at the end of an iteration.

The parallel K-means algorithm of Tian et al. [2] same strategy as the one of Zhang et al. 1. It requires the data set fully loaded on the master node and divides the data set into $m$ subset ($m$ is the number of processors). The paper only estimate the complexity of the proposed algorithm. No practical implementation is executed to make any conclusion about the empirical performance of the proposed algorithm.

A parallel K-means algorithm, called ParaKMeans [3], is implemented in the multi-threading approach on a single computer to cluster biological genes. The parallel model in this algorithm is similar as the Tian's algorithm. The only difference is ParaKMeans uses Sufficient Statistics to measure the cluster's quality and in the stop condition.

A distributed K-means algorithm is introduced in [4]. It is designed to execute on multi-processor computers. Randomly split data subset is delivered to each processor before the algorithm starts. In the beginning, each processor randomly initialize the center of its K cluster centers. In each iteration, the cluster set of each processor is re-calculated based on its data subset, broadcast its cluster set to other processors and then re-calculated its cluster set again based on the received data. The process is repeated until the cluster set is stable. In general, the parallel strategy of this paper is similar to Zhang's algorithm but without using the master node, so that it has the same similar drawbacks.

With the introduction of MapReduce in 2004 [9], all following parallel versions of K-means in this section are implemented in this framework. The MapReduce framework uses a machine to play the role of the master node in the master/slave model. The Master node splits data into subsets, sends each subset to each Mapper (playing a role of a slave node), invokes the action of all Mappers. Each Mapper executes on its data subset to create intermediate data and then sends its intermediate data to a Reducer. Reducers collect intermediate data from Mappers to create final results. A node (Master, Mapper, Reducer) can be an independent computer.

Chu's work [10] proposed a framework that applied the parallel programming method of MapReduce on several Machine Learning algorithms, including K-means. In the parallel K-means algorithm in this framework, each Mapper works on different split data and then return its intermediate data (the sum of vectors in each data subgroup) to the Master node. After collecting all intermediate data from Mappers, the Master send received data to Reducer to compute the new centroids and return the final clustering result to Master. Comparing to the original K-means algorithm, the repeated converge process is not mentioned in the parallel version in this framework, so that the parallel version cannot reach to a good clustering result.

Another parallel K-means algorithm, called PKMeans, is introduced in [11] and implemented on MapReduce. This algorithm is also used in [12] for document clustering. Another type of nodes, called Combiner, is used in PKMeans. In each iteration, the Mapper, Combiner and Reducer are serially executed. A Mapper only assigns each sample to the closest center. A Combiner, that is executed on the same computer with the Mapper, partially sum the values of the data points assigned to the same cluster and then return an array of records. Each record stores the sum of values and the number of data points of a cluster. This array will be send to the Reducer to compute the new position of cluster centers, that is a global variable and can be accessed by Mappers. Several iteration is repeated until the cluster set is stable. Therefore, the model and the limits of this work is similar to Zhang's algorithm.

In conclusion, several parallel versions of K-means uses the data parallel strategy but the parallel strategies are different when using or not using framework MapReduce. When not using MapReduce, each slave node uses the initialized data from the master node,   processes its  sub data set and then synchronizes local data by sending the master node or broadcasting to other computing nodes before repeating the next iteration. This strategy has several drawbacks, such as the master node often has to load the full data set to deliver to computing nodes, a synchronization step is required in the end of each iteration, several scans over the data set is also required. When using MapReduce, each data object is processed by Mapper so that Mapper is called several times. In addition, when the algorithm requires several iterations, communication cost between nodes is much higher.

## 3    Parallel Two-Phase K-Means (Par2PK-Means)

The new proposed algorithm, called Parallel Two-Phase K-means (Par2PK-means), is developed based on the model of 2PK-means and implemented on Hadoop following the MapReduce model. The detailed algorithm of Par2PK-means is described in Figure 1. When there is an available Mapper, it reads a data segment and executes the IKM algorithm on that data segment to create an intermediate clustering result. Reducer retrieves the intermediate clustering result of all Mappers. When all Mappers finish their tasks, Master invoke Reducer to execute IKM on all received intermediate results from Mappers to create the final clustering result.

Parameter $K_t$ decides the speed and the clustering quality of Par2K-means. Phase 1 of Par2PK-means plays a role of a compressor that reduces the number of data objects in a data buffer to $K_t$ clusters. Parameter $K_t$ decides the size of the intermediate data that enters Phase 2 so that controls the speed of the algorithm. The smaller $K_t$ is, the faster Phase 1 and Phase 2 of Par2K-means are. However, $K_t$ cannot be  too small.

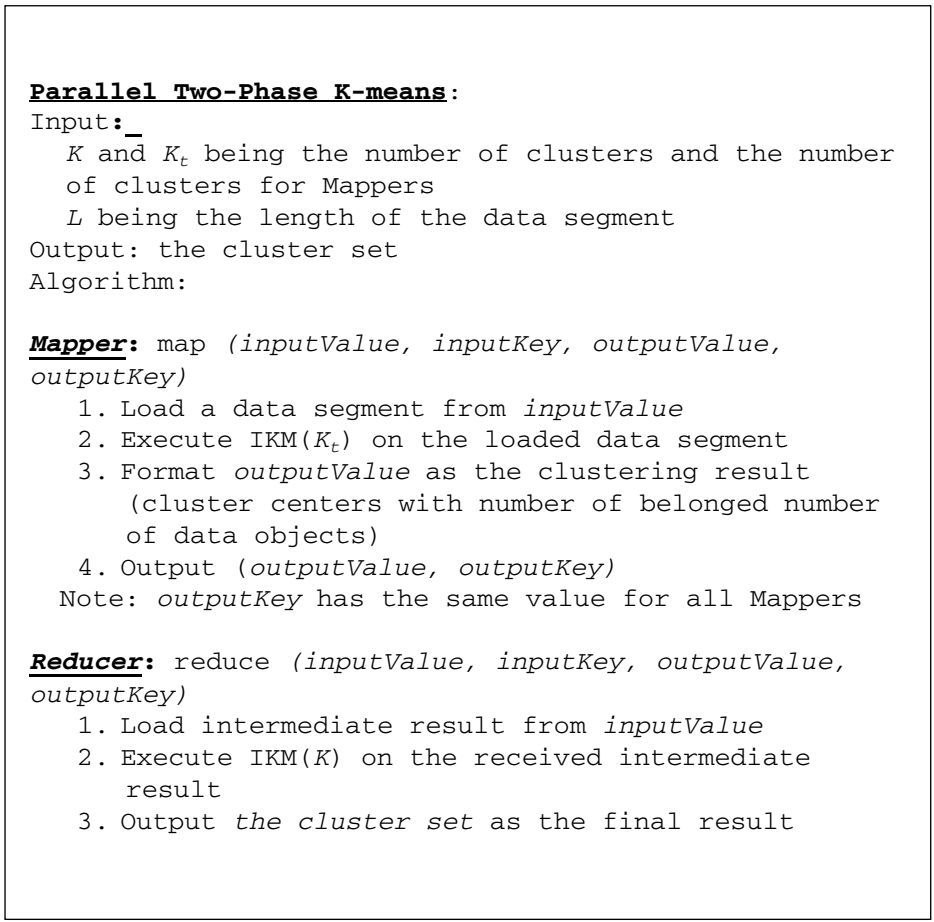

**Parallel Two-Phase K-means**:
Input:
　K and $K_t$ being the number of clusters and the number of clusters for Mappers
　L being the length of the data segment
Output: the cluster set
Algorithm:

*Mapper*: map *(inputValue, inputKey, outputValue, outputKey)*
　1. Load a data segment from *inputValue*
　2. Execute IKM($K_t$) on the loaded data segment
　3. Format *outputValue* as the clustering result
　　　(cluster centers with number of belonged number of data objects)
　4. Output (*outputValue, outputKey*)
　Note: *outputKey* has the same value for all Mappers

*Reducer*: reduce *(inputValue, inputKey, outputValue, outputKey)*
　1. Load intermediate result from *inputValue*
　2. Execute IKM(*K*) on the received intermediate result
　3. Output *the cluster set* as the final result

**Fig. 1.** The Parallel Two-Phase K-means algorithm

Par2K-means uses the model of Two-Phase K-means, so that it only produces an approximate result of the sequential version of Two-Phase K-means. The larger $K_t$ is, the higher the quality of clustering result is. Therefore, the selection of $K_t$ is trade-off between the speed and the clustering quality. In Two-Phase K-means, $K_t$ is selected correspondent to the size of a data buffer. With big data sets, $K_t$ can be selected around one thousands to one percentage of the size of data buffer without much reduction in clustering result.

## 4    Evaluation

The proposed algorithm is evaluated its speed-up ratio when comparing to its serial version. Experiments are executed on a virtual computer cluster powered by Openstack [17] which is an open source cloud computing platform. It provides mechanisms to provision the virtual machines from resource pool. To provide the virtual machine, we need to have the hypervisor which allows to create a virtual machine out of a physical one. Openstack supports several hypervisor such as Xen[16], VMWare[14], KVM[15]. In our case, we use KVM. In order to create the cluster of virtual machines, we have created the machine image that has the Hadoop version 1.0.4 and Java 1.6.0_26 installed. Hadoop consists of two sub components: Hadoop MapReduce and Hadoop Distributed File System.

**Data Distribution:** HDFS is a distributed file system based on Master/Slave model, which prepares an environment for Hadoop work. In HDFS, the master node (also called NameNode) manages all events read/write activities of the system. It splits the input data into blocks whose size is specified by user (either 64MB or 128MB) and distributes these blocks to other slave nodes called DataNodes. The master node keeps the map of data blocks and DataNodes. For fault tolerance, user can specify the number of replications of a data block on the cluster. DataNode propagates the data blocks to the specified number of nodes.

**Computing:** The Hadoop MapReduce runs on top of HDFS. Hadoop MapReduce utilizes the capability of data awareness of HDFS to distribute the appropriate computing tasks to slave nodes. It also uses the Master/Slave architecture, which contains A JobTracker(master) and a number of TaskTracker nodes (slaves). JobTracker queries the locations of data on NameNode and deliver tasks to TaskTracker nodes. The results then are aggregated and reported to the user by the JobTracker.

As we show in the next paragraphs, we use several cluster size. We use the same virtual machine configuration though (Each virtual computer has a 2.8GHz CPU and 1GB of memory).

The first experiment uses data set CoverType from the UCI Repository [13]. This data set has 581,012 data objects with 52 attributes. In this experiment, only the first 4 attributes are used. The data set is enlarged by adding noise to the original data set. The enlarged data set has 2,324,048 data objects, 4 attributes, a size of 87.6MB. The number of slave nodes are 1, 2, 4 and 8. The number of data segments is selected as 100. Number of clusters $K$ and $K_t$ are selected as 10.

The speed-up ratios of Par2PK-means execution times on different number of slave nodes comparing to its execution time on one slave node are shown in Figure 2. From the figure, the speedup ratio of Par2PK-means approaches its limit (linear speedup). The lost percentage of Par2PK-means compared to the maximum ratio is around 10% due to the initialization of Hadoop.

In the second experiment, the CoverType data set is enlarged by adding noise to the original data set. The enlarged data set has 29,050,600 data objects, 4 attributes, a size of 1.23GB. The number of segments is also selected as 1000, but the number of clusters $K$ and $K_t$ are increased to 20 due to the larger data size. The evaluation results on the speedup ratio in this experiment is shown in Figure 3. With a larger data set and bigger number of clusters, the calculation of a Mapper is much longer, so that the percentage of time of initializing tasks and Phase 2 execution on the calculating time is reduced. This reduction means the speedup ratio of Par2K-means is closer to the linear situation (a perfect situation for parallelizing an algorithm).

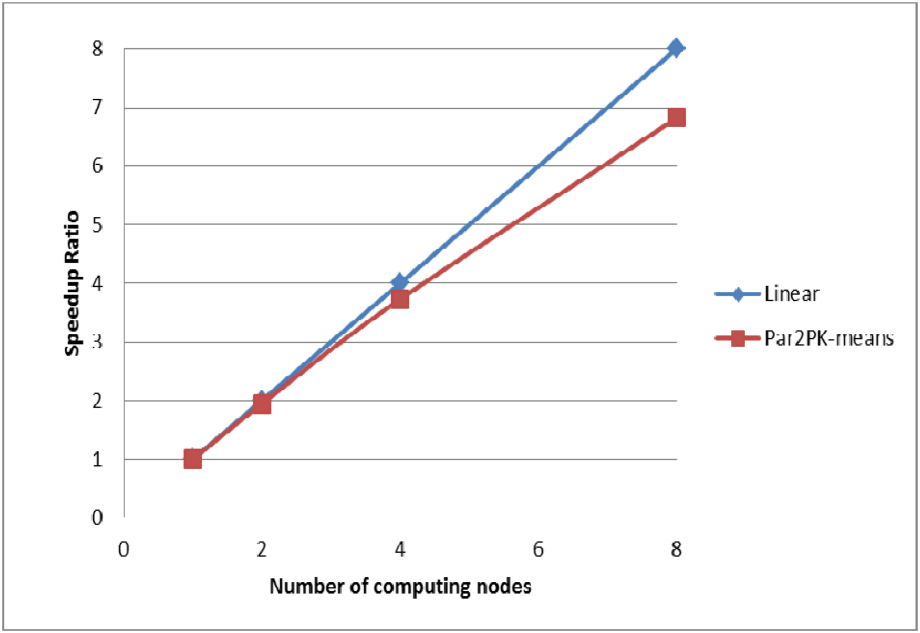

**Fig. 2.** Evaluation results on the original CoverType data set

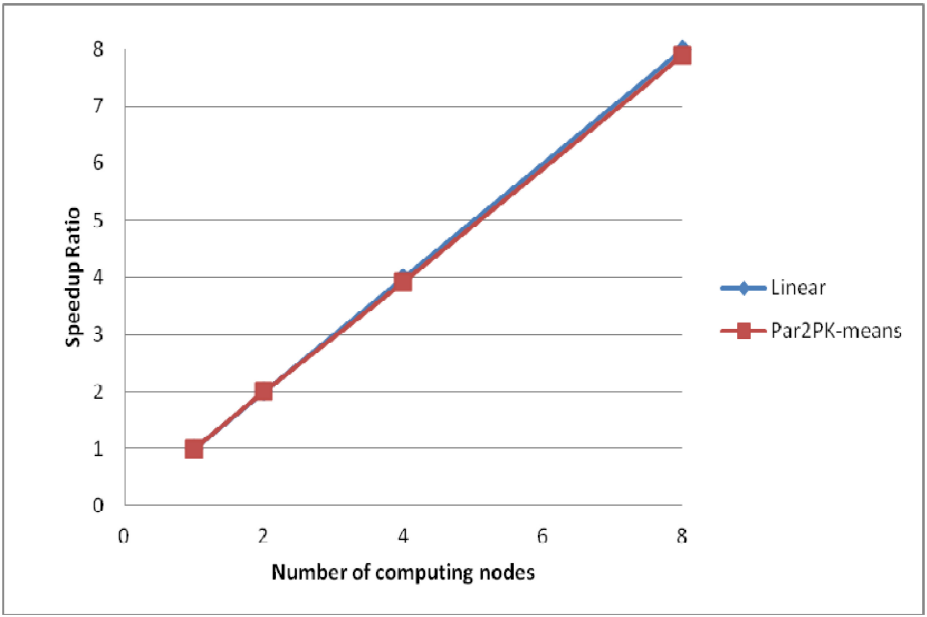

**Fig. 3.** Evaluation results on the enlarged CoverType data set

## 5   Conclusion

The paper introduces the parallel version of Two-Phase K-means. The proposed algorithm has achieved a good speedup ratio on tested data sets. However, its performance can be analyzed with other parallel strategies. Its incremental attribute (the algorithm can stopped on a number of data segments and give the best so-far clustering result) should be studied.